\documentclass{article}

\usepackage{microtype}
\usepackage{graphicx}
\usepackage{subfigure}
\usepackage{booktabs} 
\usepackage{amsfonts}

\usepackage{hyperref}

\usepackage[ruled,vlined]{algorithm2e}

\usepackage{mathtools}
\DeclarePairedDelimiter\ceil{\lceil}{\rceil}

\usepackage[accepted]{icml2021}

\icmltitlerunning{Solving Challenging Dexterous Manipulation Tasks With Trajectory Optimisation and Reinforcement Learning}

\begin{document}

\twocolumn[
\icmltitle{Solving Challenging Dexterous Manipulation Tasks With Trajectory Optimisation and Reinforcement Learning}



\icmlsetsymbol{equal}{*}

\begin{icmlauthorlist}
\icmlauthor{Henry Charlesworth}{warwick}
\icmlauthor{Giovanni Montana}{warwick}
\end{icmlauthorlist}

\icmlaffiliation{warwick}{Warwick Manufacturing Group, University of Warwick, Coventry, United Kingdom}

\icmlcorrespondingauthor{Henry Charlesworth}{H.Charlesworth.1@warwick.ac.uk}
\icmlcorrespondingauthor{Giovanni Montana}{G.Montana@warwick.ac.uk}

\icmlkeywords{Reinforcement Learning, Trajectory Optimisation, Dexterous Manipulation}

\vskip 0.3in
]



\printAffiliationsAndNotice{}  

\begin{abstract}
Training agents to autonomously control anthropomorphic robotic hands has the potential to lead to systems capable of performing a multitude of complex manipulation tasks in unstructured and uncertain environments. In this work, we first introduce a suite of challenging simulated manipulation tasks where current reinforcement learning and trajectory optimisation techniques perform poorly. These include environments where two simulated hands have to pass or throw objects between each other, as well as an environment where the agent must learn to spin a long pen between its fingers. We then introduce a simple trajectory optimisation algorithm that performs significantly better than existing methods on these environments. Finally, on the most challenging ``PenSpin" task, we combine sub-optimal demonstrations generated through trajectory optimisation with off-policy reinforcement learning, obtaining performance that far exceeds either of these approaches individually. Videos of all of our results are available at: \href{https://dexterous-manipulation.github.io/}{https://dexterous-manipulation.github.io/}.
\end{abstract}

\section{Introduction}
\label{intro}

Developing dexterous multi-fingered robotic arms capable of performing complex manipulation tasks is both highly desirable and extremely challenging. In industry today, a majority of robots make use of simple parallel jaw grippers for manipulation. Whilst this works well in structured settings, as we look to develop more autonomous robots capable of performing a wider variety of tasks in more complicated environments it will be necessary to develop considerably more sophisticated manipulators. The most sophisticated and versatile manipulator that we know of is the human hand --- capable of tasks ranging from complex grasping to writing and tool use. As such, it is natural to attempt to create robotic hands that mimic the human hand and that can perform similarly complicated manipulation tasks.

This is difficult for traditional robotic control approaches, both with real robotic hands and in simulation. The primary challenges stem from the need to precisely coordinate numerous joints as well as complex, discontinuous contact patterns that can arise at a large number of potential contact points between the hand and the object being manipulated. This motivates the use of techniques that can learn directly via interactions with the environment, without requiring accurate physics models of the hand-object system. In recent years, there has been significant success in applying both reinforcement learning (RL) and gradient-free trajectory optimisation to a range of dexterous manipulation tasks, both in simulation \cite{plapperther,dapg,polo} and with real robotic hands \cite{dactyl,openai2019solving,pddm}. Nevertheless, many of these successes involve tasks that by human standards are relatively simple, and complex manipulation tasks still remain a significant challenge.

In this paper we make three primary contributions. Firstly, we introduce a suite of challenging new manipulation tasks based on OpenAI Gym's simulated shadow-hand environments. These include tasks that require coordination between two hands, such as handing over and catching objects, as well as a challenging ``PenSpin" environment where a hand has to learn to spin a long pen between its fingers. We demonstrate that many of these tasks are extremely difficult for existing RL/trajectory optimisation techniques, making them potentially useful benchmarks for testing new algorithms.

Secondly, we introduce a simple trajectory optimisation algorithm that is able to significantly outperform existing methods that have been applied to dexterous manipulation, achieving high levels of success on most of the tasks considered.

Finally, on the environment for which our trajectory optimisation algorithm struggles the most (``PenSpin"), we introduce a method for combining the sub-optimal partial solutions generated via trajectory optimisation with off-policy reinforcement learning. We show that this leads to substantially improved performance (effectively being able to spin the pen quickly and indefinitely) --- far beyond either trajectory optimisation or RL on their own. Given that the task of spinning a pen in this way is challenging for a majority of humans, we would argue that this represents one of the most striking examples to date of a system autonomously learning to complete a complex dexterous manipulation task.

\section{Related Work}
\subsection{Dexterous Manipulation}

There is a large body of work that has looked at both designing and developing controllers for anthropomorphic robotic hands. Some work has found success by simplifying the design of the hands \cite{softhand}, however a number of trajectory optimisation/policy search methods have also been found to produce successful behaviour for more realistic robotic hands \cite{contactinvariantopt,realtimesynthesis,polo}. \citet{polo} is particularly relevant to us since their method only requires access to sampling interactions with the simulated environment and not any detailed model of the physics of the system under consideration. Here, they combine a trajectory optimisation technique known as ``model predictive path integral control" (MPPI) \cite{MPPI} with value function learning and coordinated exploration in order to perform in-hand manipulation of a cube using a simulated ADROIT robotic hand. In fact, they also demonstrate that this in-hand manipulation can also be performed only using the MPPI trajectory optimisation algorithm itself.

In terms of applying RL to anthropomorphic robotic hands, \citet{dapg} introduced a suite of dexterous manipulation environments again based on a simulated ADROIT hand, including in-hand manipulation and tool use tasks. Whilst they demonstrated that on-policy RL could be used to solve the tasks when carefully shaped reward functions were designed, they noted that this still took a long time and produced somewhat unnatural movements. They found that they could greatly improve the sample efficiency as well as learn behaviours that looked ``more human" by augmenting the RL training with a small number of demonstrations generated via motion capture.

\citet{plapperther} introduced a number of multi-goal hand manipulation tasks using a simulated Shadow hand robot and integrated these into OpenAI Gym \cite{openaigym}. These environments form the basis for the new environments we introduce in the next section. In this work they also carried out extensive experiments using Hindsight Experience Replay (HER) \cite{HER}, a state-of-the-art off-policy RL algorithm designed for multi-goal, sparse reward environments. This worked well for some of the environments, however for the more challenging ones (particularly HandManipulateBlockFull and HandManipulatePenFull) they found only limited success. A number of other works have built upon HER \cite{cer,sher}, however they only provide minor improvements on the challenging hand manipulation tasks.

\citet{pddm} introduce a model-based approach in order to solve a number of dexterous manipulation tasks in both simulation and on a real robotic hand, including manipulating two Baoding balls and a simplified handwriting task. Their method involves learning an ensemble of dynamics models and then planning each action using MPPI within these learned models, and is sample efficient enough that it can be used to learn using a real robotic hand.

\citet{dactyl} use large-scale distributed training with domain randomization to train an RL agent in simulation that can transfer and operate on a real anthropomorphic robotic hand. Here, they are able to teach the hand to perform in-hand manipulation to rotate a block to a given target orientation, and show that it is sometimes capable of achieving over 50 different goals in a row without needing to reset. \citet{openai2019solving} extend this work, introducing automatic domain randomization and training the hand to learn how to solve a Rubik's cube (more precisely, it learns how to implement the instructions given to it by a separate Rubik's cube solving algorithm). This requires considerably more dexterity than the block manipulation task, and represents one of the most impressive feats of learned dexterous manipulation to date, particularly as it is capable of operating on a real robotic hand. Nevertheless, it is worth noting that the amount of compute needed to obtain these results was enormous --- using 64 Nvidia V100 GPUs and $\sim 30,000$ CPU cores, training for several months to obtain the best Rubik's cube policy.

\subsection{Combining RL with Demonstrations}
A number of methods that make use of demonstrations in order to improve/speed up RL have been proposed. \citet{dqfd} introduce a method to combine a small number of demonstrations to speed up Deep Q-learning, sampling demonstration data more regularly and using a large margin supervised loss that pushes the values of the demonstrator's actions above the values of other actions. \citet{ddpgfd} extend this to environments with continuous action spaces, combining demonstrations with deep deterministic policy gradient \cite{ddpg}. \citet{overcomingexploration} use a similar approach but for sparse-reward, multi-goal environments. Here they combine demonstrations with HER, supplementing the RL with a behavioural cloning loss and a ``Q-filter" which ensures this only acts on expert actions with higher predicted value than the policy's proposed action. This is particularly relevant to this work, as they also consider resetting episodes to randomly chosen demonstration states to aid with exploration, which is the basis for our approach of combining demonstrations generated by our trajectory optimisation algorithm with RL. \citet{gcil} also introduce a method designed for multi-goal environments, effectively combining HER with generative adversarial imitation learning \cite{gail}.

\section{New Dexterous Manipulation Tasks}
\begin{figure*}
    \centering
    \includegraphics[width=1.0\linewidth]{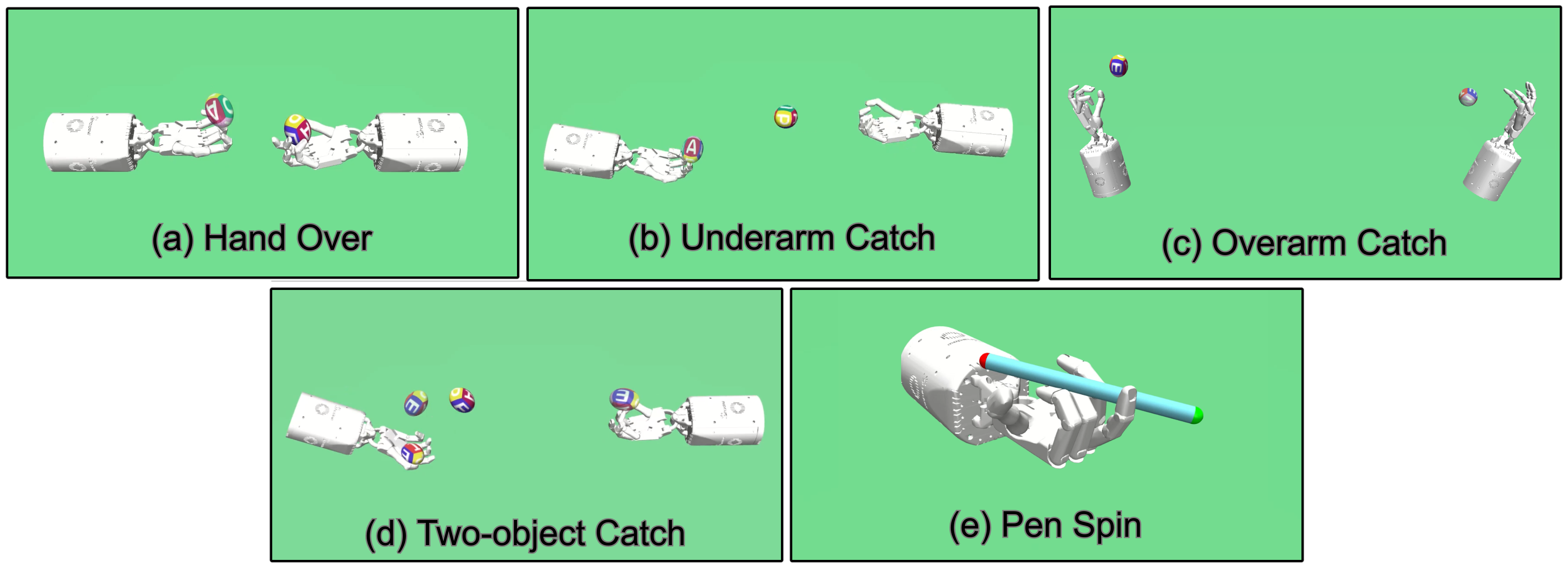}
    \caption{Overview of the types of new environments we introduce in ``dexterous-gym". For (a)-(d) transparent objects visualise the desired goals, whilst opaque objects are the physical objects being manipulated.}
    \label{fig:newenvs}
\end{figure*}
In this section we introduce ``Dexterous-gym" --- a suite of challenging hand-manipulation environments based on modifications to the openAI gym shadow hand environments \cite{plapperther}\footnote{All of our new environments are open-sourced at: \href{https://github.com/henrycharlesworth/dexterous-gym}{ https://github.com/henrycharlesworth/dexterous-gym}}. These include a number of environments that require coordination between two hands, such as handing over an object and playing catch with either one or two objects. This means that their state/action spaces are at least twice as large as the standard OpenAI gym manipulation tasks. All of the two-handed environments are goal-based in the same way as the original environments, meaning that a new goal is generated for each episode, and each environment is available with both sparse and dense reward variants (sparse rewards given only when the desired goal is achieved, vs. a dense reward based on a measure of distance to the desired goal). For each, the object can be either an egg, block or a pen.

We also introduce an environment we call ``PenSpin", which is the only non goal-based, single-hand environment. This is a simple modification of the HandManipulatePen environment from OpenAI Gym where we re-define the reward function to encourage the pen to be spun around (whilst remaining horizontal relative to the hand). The basic types of environment introduced are shown in Figure \ref{fig:newenvs}.

\textbf{(a) Hand Over environments:} In these environments we have two hands with fixed bases slightly separated from each other. The goal position/orientation of the object will always only be reachable by the hand that does not start with the object, so to achieve the goal the object must first be passed to the other hand (e.g. by flicking it across the gap). Observation space: 109-dimensional, action space: 40-dimensional, goal space: 7 dimensional.

\textbf{(b) Underarm Catch environments:} Similar to the Hand Over environments, except now the bases are not fixed and have translational and rotational degrees of freedom that allow them to move within some range. However, the hands are too far apart to directly pass over the object and so it must be thrown and subsequently caught by the other hand, before being moved to the desired goal. Observation space: 133-dimensional, action space: 52-dimensional, goal space: 7-dimensional.

\textbf{(c) Overarm Catch environments:} Similar to Underarm Catch, except the hands are oriented vertically, requiring a different throwing/catching technique. Observation space: 133-dimensional, action space: 52-dimensional, goal space: 7-dimensional.

\textbf{(d) Two-object Catch environments:} Similar to Underarm Catch but now each hand starts with its own object. The goal is then the position/orientation of both objects, which are always only reachable by the other hands. This means that both objects have to be thrown in order to be swapped over. This requires significantly more coordination than the single-object variant. Observation space: 146-dimensional, action space: 52-dimensional, goal space: 14-dimensional.

\textbf{(e) Pen Spin environment:} The set-up is completely identical to HandManipulatePen from OpenAI Gym, with a single hand interacting with a long pen, however we remove the notion of a goal and instead define a new reward function; $r = \omega_3 - 15 |b_z-t_z|$, where $\omega_3$ is the third-component of the pen's angular velocity (in the generalised coordinates used by Mujoco) and $b_z$ and $t_z$ are the z-positions of each end of the pen. The first term encourages the pen to be spun, whilst the second penalises the pen for becoming vertical. Although lower-dimensional than the other tasks, this environment requires a significant amount of dexterity, and indeed most humans struggle to perform a similar task successfully. Observation space: 61-dimensional, action space: 20-dimensional.

\section{Methods}
\subsection{Trajectory Optimisation for Precise Dexterous Manipulation (TOPDM)}

In this section we introduce TOPDM, a trajectory optimisation algorithm that we apply to both the environments introduced in the previous section as well as the most challenging OpenAI gym manipulation tasks. Our method shares a number of similarities with MPPI \cite{MPPI}, and so we begin by describing that in detail before going on to highlight the key differences with our approach \footnote{The code for all of our experiments is available at: \href{https://github.com/henrycharlesworth/SCDM}{https://github.com/henrycharlesworth/SCDM}}.

MPPI is a gradient-free trajectory optimisation technique that can plan actions purely through sampling trajectories, either with a simulator (as in \citet{polo}) or using a learned model (as in \citet{pddm}). It is a form of model-predictive control where the next action is chosen by planning over a finite number of future time steps, $\tau$, before taking a single action and then re-planning again from the next state. The procedure is quite simple --- to plan the next action from the current state $s_t$, $N$ mean action sequences are initialised with random noise: $\{ \mu_{t:t+\tau}^{(n)} \}_{n=1}^N$. Rather than executing these mean actions directly, a coupling between subsequent actions is introduced such that the actual actions executed are $a_{t'}^{(n)} = \beta \mu_{t'}^{(n)} + (1-\beta) a_{t'-1}^{(n)}$, where $\beta \in [0,1]$. The motivation for this is that an intermediate $\beta$ value can produce smoother action sequences and effectively reduces the dimensionality of the search space (by excluding non-smooth action sequences). These action sequences are then all evaluated using the simulator (or model) and their returns over the $\tau$ time steps considered, $R_n$, are recorded. These are then used to update the mean action sequence by calculating $\mu_{t:t+\tau} = \frac{\sum_{n=1}^N \mu_{t:t+\tau}^{(n)} e^{\kappa R_n}}{\sum_{k=1}^N e^{\kappa R_k}}$, i.e. an exponentially weighted average of all of the mean action sequences weighted by their returns (with $\kappa$ as a hyperparameter). If this is the final iteration, we return $a_t = \beta \mu_t + (1-\beta) a_{t-1}$ as the action to actually execute in the environment, otherwise we duplicate $\mu_{t:t+\tau}$ $N$ times, add noise, and repeat the procedure.

The trajectory optimisation algorithm we implement shares the same basic structure as MPPI, but we make three key changes:
\begin{enumerate}
    \item Firstly, rather than taking a weighted average of action sequences we instead choose some fraction $f_b$ of the best performing trajectories at the end of an interation and uniformly duplicate them so that there are $N$ in total. These get carried over as the initialisation of the next iteration of the plan (on the final iteration we return the first action from the single best performing trajectory we have found throughout the full planning process). This is similar to the ``cross-entropy method" used for trajectory optimisation in \citet{planet}, except there they choose the top $f_b$ fraction of trajectories in order to update the parameters (mean and standard deviation) of a normal distribution that is used to sample the action trajectories for the next iteration. The motivation for doing away with the averaging procedure is that when we are considering high-dimensional tasks that require significant precision, it might be very rare to see a trajectory that significantly improves upon the return. This means that any averaging procedure can easily perturb or effectively ignore any trajectories that have unusually high returns and significantly reduce overall performance.
    
    \item Secondly, rather than starting each plan from scratch, we instead carry over the best trajectory from the previous planning step and use this to initialise the mean action sequence. That is, given the best mean action sequence from the previous planning step, $\mu_{t:t+\tau}^b$, we use $\mu_t^b$ to choose the next action actually taken in the environment, but rather than discarding $\mu_{t+1:t+\tau}^b$ we instead use it to initialise the first $\tau-1$ actions for all of the mean action sequences for the next planning step. The reason for not doing this in MPPI is motivated as wanting to encourage exploration and not getting stuck in local optima. Whilst this is a legitimate concern and may be important in some situations, we find that for the challenging tasks considered in this paper it does not cause any issues. Indeed, since the problems considered require such precise coordination we find it is much more useful to make use of as many iterations of improvement as possible (and carrying over the previous best mean sequence effectively means building upon all of the iterations carried out in the previous plan, rather than restarting the search from scratch).
    
    \item Finally, we change the way in which we add noise to the mean action sequences. At the start of a given iteration of planning we can consider the collection of mean action sequences as a multi-dimensional array, $\mu \in \mathbb{R}^{N \times \tau \times a_d}$, where $a_d$ is the dimension of a single action. Rather than adding noise to this whole vector, for a given iteration we instead choose to add noise only to a fraction $f_n$ of randomly chosen dimensions. The motivation here is that if we have already found a reasonably good trajectory then adding noise to all dimensions of all time-steps can very easily perturb the trajectory away from good performance. If we only add noise to some of the dimensions at some time steps we make it more likely that we can find slight improvements to certain parts of the trajectory without disturbing the rest of the sequence. As an analogy, consider mutations to a DNA sequence. If at each iteration every member of the population had its whole sequence mutated this would make incremental improvement much more difficult --- whereas many iterations of a small number of mutations to each member of the population allows for useful mutations to accumulate.
    
\end{enumerate}

The full details of this method are laid out in Algorithm 1. In section 5.3, we study the effect of each of these modifications individually on a subset of the environments considered to get an idea of which has the largest impact on performance.

\begin{algorithm*}
\DontPrintSemicolon
\textbf{Initialise:} $\tau$ (planning horizon), $N_t$ (number of trajectories per iteration), $N_i$ (number of iterations), $T$ (trajectory length), $\beta$ (action coupling parameter), $f_n$ (fraction of non-zero noise-dimensions), $f_b$ (fraction of best trajectories retained for next iteration), $\sigma_n$ (noise standard deviation), $\sigma_i$ (initial standard deviation), $a_d$ (action-dimension), $E$ (environment)

\Begin{
Initialise $\mu \in \mathbb{R}^{N \times \tau \times a_d} \sim \mathcal{N}(0, \sigma_i)$, $a \in \mathbb{R}^{N \times \tau+1 \times a_d}$ as zeroes \;

\For{$t=1:T$}{

\For{$i=1:N_i$}{
add noise $\sigma_n$ to $\ceil{f_n \times a_d}$ randomly chosen dimensions of $\mu$ \;

\For{$s=1:\tau$}{
$a[:, s, :] = \beta \ \mu[:, s, :] + (1-\beta) \ a[:, s-1, :]$\;
}
Evaluate all $N_t$ of these action trajectories with $E$. Sort based on returns \;

Take fraction $f_b$ of best performing trajectories. Duplicate these to uniformly fill $\mu$ and $a$ with copies of these best trajectories. These act as the starting point for the next iteration \;
}

Take $\mu_b$ as the best performing mean action sequence from any iteration. \;

Step real environment E with $a_t = \beta \mu_b[1] + (1-\beta) a_{t-1}$ \;

$\mu[:, 1:\tau-1, :] = \mu_b[2:\tau, :]$ (carry over rest of action sequences for next search) \;

$a[:, 0, :] = a_t$ (set previous action)\;
}

}
\caption{Trajectory Optimisation for Precise Dexterous Manipulation (TOPDM)}
\end{algorithm*}

\subsection{Using Demonstrations Gathered via Trajectory Optimisation to Guide RL}
There are two main issues with the trajectory optimisation approach described in the previous section. Firstly, despite being highly parallelisable, for the challenging manipulation environments we consider in this paper the algorithm still requires a lot of compute in order to generate high quality trajectories. On the other hand, a neural network may take a long time to be trained with RL, however once it has been trained it can easily generate new trajectories on the fly in real time. The second issue with the trajectory optimisation approach is that it is necessary to plan over a finite horizon, $\tau$. Previous work \cite{polo} has looked at overcoming this by concurrently learning a value function whilst generating experience via trajectory optimisation, however for our case where each planning step is so expensive this becomes more challenging. We find that in our dexterous manipulation experiments it is not practical to extend $\tau$ much beyond $\sim25$ (although for the humanoid experiments we do extend to $\tau=40$), which is fine for many of the environments considered but can lead to situations where optimising reward over a relatively short period of time misses strategies that over a longer time-frame can lead to much higher rewards. As we shall see in the next section, this turns out to be the case for the PenSpin environment. This is another drawback that RL does not suffer from, as with RL we aim to maximise the (usually discounted) sum of future rewards, which in principle allows us to learn strategies that maximise rewards over a much longer period of time.

Both of these issues motivate combining the demonstrations generated via trajectory optimisation with RL. We focus on the PenSpin environment for two reasons --- firstly because it is the only ``standard" RL environment (i.e. non goal-based), but mainly because it is the environment which requires the most dexterity and which is the most challenging for TOPDM.

The approach we take is extremely simple --- firstly we run TOPDM as described in the previous section and gather a small set of (potentially sub-optimal) example trajectories. We train an agent using the off-policy RL algorithm TD3 \cite{td3}. However, rather than gathering data by resetting the environment and running an episode, we instead consider gathering segments of data. With some probability a segment will come from a normal ongoing episode, but with some probability we will instead load the simulator to a randomly chosen demonstration state and gather the segment of data from there instead. We start with a high probability of gathering data starting from a demonstration state and then gradually anneal this throughout the training.

The motivation for this approach is that it allows us to gather lots of data that start from desirable configurations where we know it is possible to obtain high-rewards, as well as states that we know can lead to these desirable configurations. If we consider that for a particular task there may be states that are potentially highly rewarding but which are very difficult to discover through random exploration then there are essentially two challenges. Firstly, the agent needs to be able to discover these states at all. But even if the agent can eventually discover the states it needs to be able to estimate their values. If an agent finally discovers a hard to reach state but has no concept of its value, then it will not necessarily know that it wants to return there again later and will most likely not discover it again for a long time. The approach we propose here therefore helps with exploration in two ways --- firstly, it allows the agent to learn better value estimates of states that are potentially very difficult to reach via random exploration much earlier on, meaning that when an agent does eventually discover them it will know to return. Secondly, it also helps with learning to reach these states by sometimes starting the simulation from states that are much closer to the desirable states than the initial state at the start of a normal episode.

We also employ the same action-coupling technique used in the trajectory optimisation during the training of the RL agent. That is, the agent's policy takes in the current observation $o_t$ as well as the previous action, $a_{t-1}$, and outputs $\mu_{\theta}(o_t, a_{t-1})$ ($\theta$ represent the parameters of the policy). The actual action taken in the environment is then $\beta \mu_{\theta}(o_t, a_{t-1}) + (1-\beta) a_{t-1}$. Interestingly, in Appendix B we provide evidence that this step is crucial to reliably train an agent to solve this task.

Finally, it is worth noting that similar ideas have been explored before, but in different contexts. \citet{overcomingexploration} include resetting to observed states when using demonstrations to aid with exploration in multi-goal RL, however it is more of a detail that is sometimes found to be useful and not at the core of their method (which mixes behavioural cloning and hindsight experience replay) --- and their demonstrations have to be generated manually in virtual reality. GoExplore \cite{goexplore} is another method which uses a similar principle and gathers data by resetting the simulator to promising states that it has saved in an archive. However, this requires defining some kind of similarity measure between states so that they can be distinguished and stored in a finite archive, whilst with our approach we already have small finite set of demonstrations generated through trajectory optimisation which we can choose from randomly when we choose to load the simulator to a demonstration state. Our method is the first approach that we're aware of that combines demonstrations autonomously generated via trajectory optimisation with RL.

\section{Results}
\subsection{OpenAI Gym Manipulation Environments}
\begin{figure}
    \centering
    \includegraphics[width=1.0\linewidth]{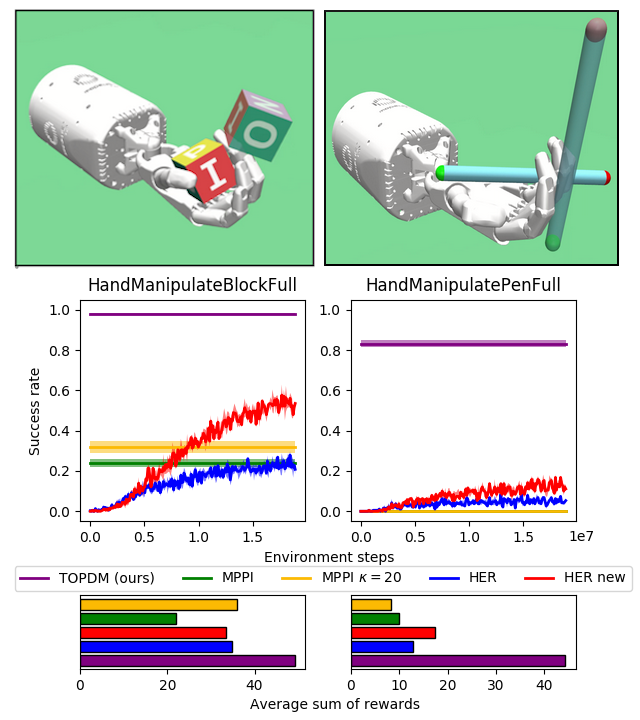}
    \caption{Performance of TOPDM compared with various baselines on the two most challenging dexterous manipulation environments from OpenAI Gym. The top two panels show screenshots of the two environments considered, with the transparent objects representing the target object goals. The plots show the performance in terms of both success rate and average rewards obtained in an episode for our method and a number of baselines. Note that in the success rate plots we have a slight difficulty in that we are comparing learning-based approaches (HER) with trajectory optimisation approaches. The environment steps label in this figure (and in Figures 3 and 5) apply only to the learning-based approaches, and we do not want to give the impression that the trajectory optimisation approaches require significantly less interactions in order to achieve their performance (indeed, they require substantially more interactions overall). Full details of how many interactions are required for each trajectory generated with TOPDM (and MPPI) are included in Appendix A.}
    \label{fig:gym_envs_res}
\end{figure}

We start by evaluating our method on the two most challenging OpenAI Gym manipulation tasks --- HandManipulateBlockFull and HandManipulatePenFull. These both require manipulating an object to a desired orientation \textit{and} desired position simultaneously, rather than only a desired orientation. HandManipulatePenFull is particularly challenging as it is easy for the pen to get stuck in between the fingers. Figure \ref{fig:gym_envs_res} shows the results of our trajectory optimisation algorithm alongside a number of baselines. For each environment we plot both the final success rate (the fraction of episodes that end with the desired goal being achieved) as well as the average sum of rewards obtained in an episode. This latter measure is useful to include as well since it allows us to evaluate how close a method might get the goal even if it doesn't successfully achieve it at the last time step. There are a few things here worth noting: firstly, we slightly modify the reward from the default environments so that they lie in the range $[0,1]$ for each time step. Secondly, we change the maximum number of time steps down from the default of 100 to 75 as this does not significantly effect the success rate of any of the methods considered. Thirdly, we note that Hindsight Experience Replay (HER) actually works better with sparse rewards, and so for this comparison we train using the sparse-reward version of the environment but still evaluate the average sum of rewards using the dense version once a policy is trained. Finally, we note that by default the HandManipulatePenFull environment has a distance tolerance of 0.05 (i.e. the centre of mass of the pen must be within this distance of the desired position before the positional part of the goal is considered achieved), whilst the other environments have this set to 0.01. This means that the goal can be considered achieved even though visually it is clear that the pen does not even overlap at all with the desired goal. We change this value back to 0.01 in order to match up with the other environments (making the task significantly harder, which explains why the HER result is lower than previously reported in \citet{plapperther}).

For all of the trajectory optimisation experiments in this section we use $\tau=20$, $N_i=40$ and $N_t=1000$ (full hyperparameters are detailed in Appendix A). This means that this approach, whilst highly parallelisable, is computationally intensive and trajectories have to be computed offline.

We also compare our trajectory optimisation method against MPPI, which is the most directly comparable approach and has been used before in various dexterous manipulation tasks. We consider two variants, firstly setting $\kappa=1$ (as in \citet{polo}) which we just call ``MPPI", and secondly setting $\kappa=20$ (which was used for some experiments in \citet{pddm}), which we label ``MPPI $\kappa=20$". Increasing $\kappa$ changes the weight we give to trajectories, with higher values providing more weight to the highest performing trajectories. Anecdotally, we find that increasing $\kappa$ improves performance on all tasks considered, up until a point beyond which there is no further improvement (e.g. there is very little difference in performance between $\kappa=10$ and $\kappa=20$). To give the fairest comparison possible, for both of these we scale up the number of iterations as well as the number of trajectories sampled per iteration such that they are the same as used in our trajectory optimisation method.

Interestingly, when running the HER experiments we found that by changing one of the hyperparameters away from the value used in the original paper we significantly improved its performance on HandManipulateBlockFull (we changed the probability of a random action from 0.3 to 0). We include results with both the original hyperparameters (``HER") as well as with the improved values (``HER new").

We see that our method substantially outperforms all of the baselines, both in terms of final success rate as well with the average sum of rewards, achieving close to 100\% success rate of HandManipuilateBlockFull. The results on the more challenging HandManipulatePenFull are even more striking, as none of the other methods are able to achieve more than $\sim$15\% success rate whilst ours achieves $\sim$83\% along with a much higher average sum of rewards. Side-by-side videos of the final performance of all of these methods can be found in the supplementary files.

To demonstrate that the improvements we obtain over MPPI are not just a consequence of looking at dexterous manipulation tasks, in Appendix C we also run experiments on the Humanoid environment from OpenAI Gym\cite{openaigym}. We find that we are able to produce extremely high quality trajectories (often scoring $> 15,000$), significantly outperforming state of the art reinforcement learning approaches.

\subsection{Results on Our Custom Goal-based Environments}
\enlargethispage{2\baselineskip}
\begin{figure}
    \centering
    \includegraphics[width=1.0\linewidth]{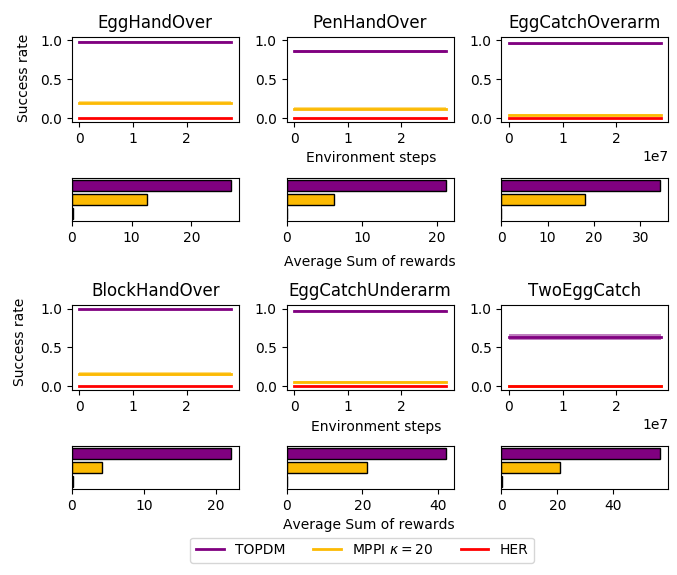}
    \caption{TOPDM and baseline performance on goal-based environments introduced in Section 3.}
    \label{fig:custom_envs_res}
\end{figure}
Figure \ref{fig:custom_envs_res} shows the results of our trajectory optimisation algorithm on some of the goal-based variants of the environments introduced in section 3. We see that all of the environments are very difficult for HER, where effectively the success rate is zero and very little reward is gathered. MPPI performs better, generally making a reasonable attempt at completing the task but nevertheless lacking the required precision to reliably achieve \textit{and maintain} the goals until the end of the episode. On the other hand, our method achieves $\sim 100\%$ success rate on four of the environments, and then $\sim 80\%$ and $\sim 60\%$ respectively on the more challenging PenHandOver and TwoEggCatch environments respectively. Note that we omit the results for MPPI $\kappa=1$ and HER original, as these both achieve practically zero success rate and close to zero sum of rewards for each of the six environments considered here.

\subsection{Ablation study: which modification to MPPI has most impact?}
\begin{figure}[!htb]
    \centering
    \includegraphics[width=0.9\linewidth]{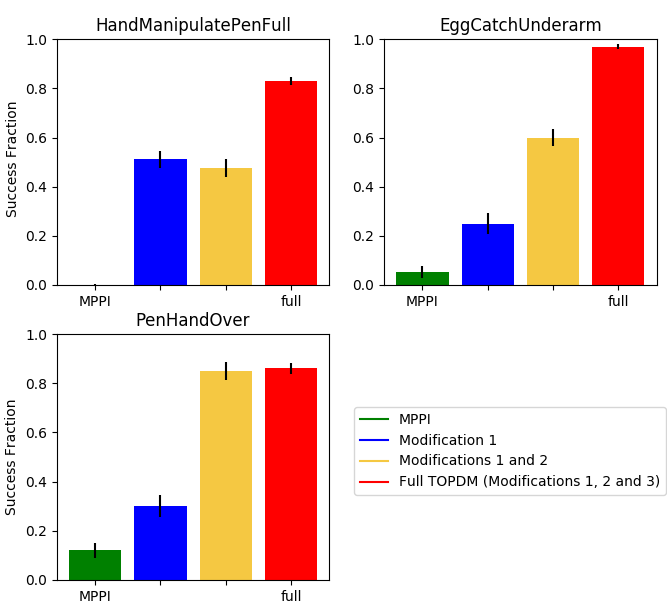}
    \caption{Trajectory optimisation ablation results.}
    \label{fig:ablation}
\end{figure}
In this section we study which of the three key adjustments we made to MPPI introduced in Section 4.1 have the largest impact. We take three of the environments considered and run additional experiments for each, firstly with just modification 1 (duplicating the best trajectories at each iteration rather than a weighted average), and then with modifications 1 and 2 (carrying over the best trajectory from the previous planning step to initialise the next step of planning). In Figure \ref{fig:ablation} we compare these results with both MPPI and the full TOPDM (modifications 1, 2 and 3). We see that the impact depends somewhat on the environment under consideration. In all three environments the first modification provides a reasonable boost in performance over MPPI alone, and the overall performance with all three modifications is always highest. We see that the second modification can have a large impact on performance for some environments, but on HandManipulatePenFull does not lead to any kind of boost (and potentially degrades performance slightly). This is likely because carrying over the previous trajectory from the previous planning iteration makes the system more prone to getting stuck in local optima. As such, there may often be situations where this modification is less desirable than the other two.

\subsection{Solving the PenSpin environment}
\begin{figure}
    \centering
    \includegraphics[width=0.9\linewidth]{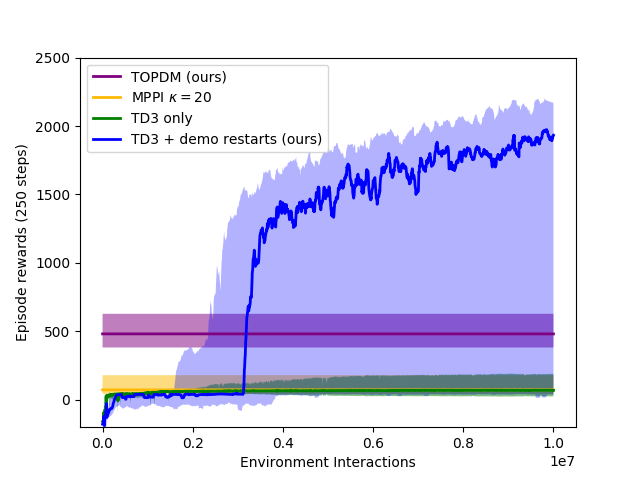}
    \caption{Results on the PenSpin environment. The thick lines represent the median performances over 10 trials, with the confidence intervals given as the minimum and maximum performing runs. Note that we include these large confidence intervals for TD3 + demo restarts because a small fraction of runs never improve significantly during training (see Appendix B for more discussion).}
    \label{fig:penspin}
\end{figure}

Figure \ref{fig:penspin} summarises the results on arguably the most challenging environment --- PenSpin. Here we see that both MPPI and TD3 (a state of the art off-policy RL algorithm) both achieve very low scores and are never able to get the pen to spin properly at all. This is probably because the environment requires a fairly precise set of actions at the start in order to move the pen into a good position to start spinning it, and finding this consistently through random exploration is very unlikely. On the other hand, our trajectory optimisation algorithm (TOPDM) performs significantly better, and almost always starts spinning the pen for some period of time. However, in none of the trials gathered was the agent able to continuously spin the pen for the whole 250 time steps without either dropping it or getting it stuck between its fingers at some point. 

Nevertheless, the imperfect demonstrations generated by TOPDM can be combined with TD3 as described at the end of section 4 (TD3 + demo restarts in Figure \ref{fig:penspin}). We see that this combination significantly outperforms both TOPDM and TD3 by themselves. Note that we plot the median return over 10 runs, with confidence intervals based on the maximum and minimum values. As two of the runs did not learn to improve beyond the base TD3 performance (the other 8 all did), the confidence interval here is large. In Appendix B we plot each trajectory individually when comparing $\beta=0.7$ vs $\beta=1.0$. 

This is a nice demonstration of how imperfect demonstrations (generated autonomously with trajectory optimisation) can be combined with RL in order to generate significantly improved performance on a challenging environment. We also note that the final trained policy can be run beyond 250 time steps (the standard episode length) and appears to be robust enough to continue to spin the pen indefinitely. This can be seen from watching the video in the supplementary files or on the project website.

In Appendix B we investigate the importance of using the action-coupling parameter $\beta$. Surprisingly, we find that not using any action coupling ($\beta=1.0$) massively reduces performance, with only 1/10 trials achieving significantly better performance than TD3 on its own, compared with 8/10 trajectories with $\beta=0.7$. Whilst this trick has been used previously with trajectory optimisation, we are not aware of it being used in RL before, and these results suggest that action-coupling could prove to be generally useful when using RL for high-dimensional continuous control tasks.

\section{Conclusion}
\enlargethispage{2\baselineskip}
We have introduced a suite of challenging dexterous manipulation tasks on which current reinforcement learning/ trajectory optimisation algorithms fail to achieve good performance. We then introduced a new trajectory optimisation technique that performs significantly better, obtaining impressive results on all of the environments considered. However, this comes at the cost of computational expense, requiring the solutions to be computed offline. We went on to show how the solutions generated by this trajectory optimisation algorithm could be used to guide the training of an RL agent which, once trained, can be deployed in real time. Applying this to the task of learning to spin a pen between the fingers of a robotic hand leads to performance that far exceeds either trajectory optimisation or RL on its own. We would argue that these results on this challenging tasks represent one of the most impressive examples to date of autonomously learning to solve a complex dexterous manipulation task.

In the future we would like to combine the trajectories generated by TOPDM with imitation learning (IL)/ RL to train an agents that can solve the goal-based tasks we introduced as well. Whilst not our primary focus, in some preliminary experiments we found that existing methods \cite{gcil, overcomingexploration} did not work well here. We also tried implementing a similar technique as with solving the PenSpin environment, but using HER rather than TD3 as the base RL algorithm. Whilst this performed slightly better than HER alone we were unable to attain performance close to a similar level of TOPDM. As such, as part of the code release we also include the trajectories generated for each task by TOPDM, and leave combining these with RL/IL as a challenge for future research. The environments introduced here also represent a significant challenge for future approaches based purely on RL.

\bibliography{example_paper}
\bibliographystyle{icml2021}

\clearpage

\appendix

\section{Experimental details}
\subsection{Trajectory optimisation (TOPDM)}
The default hyperparameters used for the trajectory optimisation experiments are the following:
\begin{itemize}
    \item $\tau = 20$ (planning horizon)
    \item $N_i = 20$ (number of iterations for each planning step)
    \item $N_t = 1000$ (number of trajectories of length $\tau$ generated for each iteration of the planning step)
    \item $f_b = 0.05$ (fraction of highest performing trajectories carried over at each iteration)
    \item $\beta = 0.7$ (action-coupling parameter)
    \item $\sigma_i = 0.9$ (initialisation noise)
    \item $\sigma_n = 0.3$ (standard deviation of noise added at each iteration)
    \item $f_n = 0.3$ (fraction of action dimensions that noise is added to)
\end{itemize}

An episode for the OpenAI Gym environments is taken to be 75 time steps. For our custom goal-based environments we take an episode to be 50 time steps, and for PenSpin we take a standard episode to be 250 time steps. For TwoEggCatchUnderarm we use $N_i=80$, $N_t=2000$ and for PenSpin we use $N_i=80$, $N_t=4000$.

\subsection{TD3 plus demos}
In our experiments with TD3 plus demonstrations generated by TOPDM we used largely standard parameters as reported in Fujimoto et al. (2018) for the core TD3 algorithm. 
\begin{itemize}
    \item start timesteps: 25,000 (initial steps gathered with random policy)
    \item total timesteps: 10,000,000
    \item exploration noise: 0.1
    \item batch size: 256
    \item $\gamma: 0.98$ (discount factor)
    \item target network update rates: 0.005
    \item policy noise: 0.2
    \item target policy noise clip threshold: 0.5
    \item update policy frequency: 2 (policy is updated every 2 critic updates)
    \item $\beta:$ 0.7 (action coupling parameter)
    \item segment length: 15 (number of timesteps per segment)
    \item initial probability of starting segment from demo reset: 0.7
    \item decay factor: 0.999996 (each step probability of demo reset multiplied by this)
\end{itemize}

Neural networks for the policy/critic and target policy/target critic were all fully connected with two hidden layers of size 256, and use the RelU activation function.

\section{Effect of action-coupling with RL plus demos for PenSpin}

\begin{figure}[!htb]
    \centering
    \includegraphics[width=1.0\linewidth]{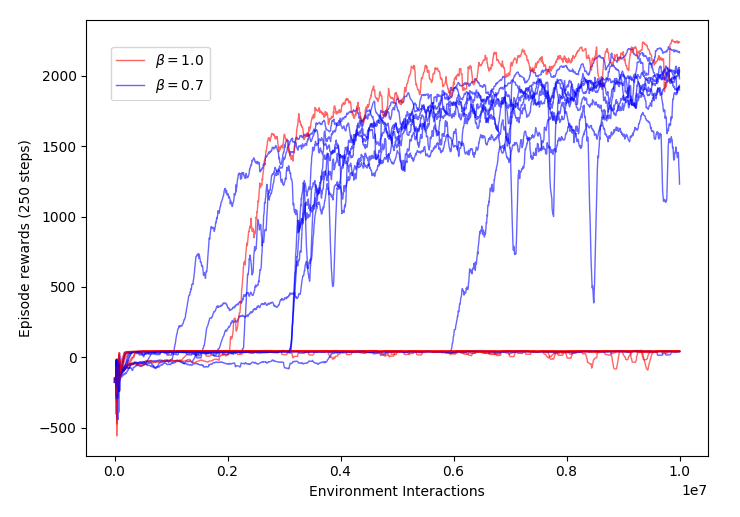}
    \caption{10 runs for $\beta=1.0$ (no action coupling) vs $\beta=0.7$ (action coupling)}
    \label{fig:penspinablation}
\end{figure}
Here we investigate the effect of including action-coupling when training the RL agent with demonstration resets on the PenSpin environment. We run 10 experiments with different initial seeds for two values of the coupling parameter, $\beta=1.0$ (no action-coupling) and $\beta=0.7$ (the value used in the results shown in Figure 4). We see that including the action-coupling allows for the agent to learn a high-quality policy significantly more often, whilst only one run with no action-coupling ($\beta=1.0$) succeeds at the task. Nevertheless, even with the action-coupling term not every run is successful, and sometimes the agent is not able to learn to get over the initial hurdle of getting the pen starting to spin. This explains why the uncertainty estimates in Figure 5 are so large (as we use the minimum and maximum scores to plot this).

\section{Experiments on the Humanoid Environment}
Our primary focus in this paper has of course been solving dexterous manipulation tasks. However, the trajectory optimisation algorithm we introduced (TOPDM) is completely and as such should perform well in other challenging environments too. To verify this we ran some experiments on the Humanoid-v3 environment in OpenAI gym. We kept the hyperparameters almost unchanged from the standard dexterous manipulation tasks, with the exception of $\tau$ which we found we had to increase to 40 (otherwise the agent could be short-sighted - learning a plan where it moves very fast over a short period of time but at the cost of becoming unstable and eventually falling over). 

\begin{figure}[!htb]
    \centering
    \includegraphics[width=1.0\linewidth]{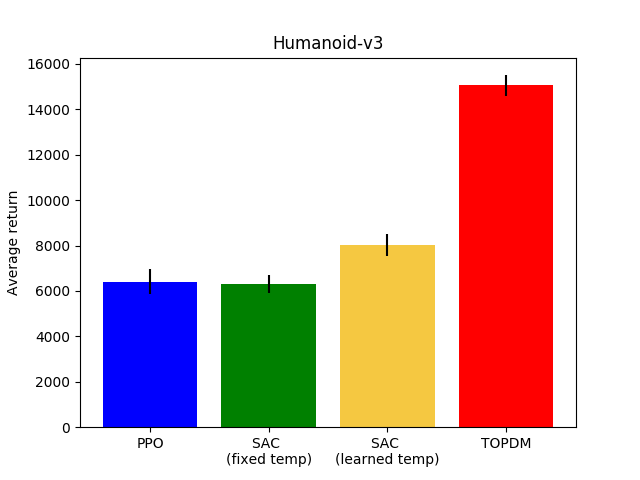}
    \caption{Our trajectory optimisation approach (TOPDM) applied to an environment not involving dexterous manipulation (Humanoid-v3). We compare the score with PPO, standard SAC and SAC with a learned temperature \cite{haarnoja2019soft}, all evaluated after 10 million steps.}
    \label{fig:penspinablation}
\end{figure}

Figure 7 shows the results on this environment compared to current state of the art RL methods. We see that TOPDM achieves a score almost double that of the best RL method, demonstrating that we can obtain extremely high quality trajectories with this approach. Watching the video of the behaviour the agent learns is also interesting, as we see the agent does not learn to run in what we would consider a normal way, but rather twists as it propels itself forwards at a very high speed.

\end{document}